\documentclass{article} \usepackage{iclr2027_conference,times}

\usepackage{amsmath,amsfonts,bm}

\def\eqref#1{equation~\ref{#1}}

\def\1{\bm{1}}

\DeclareMathAlphabet{\mathsfit}{\encodingdefault}{\sfdefault}{m}{sl}
\SetMathAlphabet{\mathsfit}{bold}{\encodingdefault}{\sfdefault}{bx}{n}

\usepackage{hyperref}
\usepackage{url}
\usepackage{graphicx}
\usepackage{booktabs}
\usepackage{multirow}
\usepackage{xcolor}
\usepackage{wrapfig}
\usepackage{pifont}
\usepackage{algorithm}
\usepackage{algpseudocode}

\newcommand{\best}[1]{\textcolor{red}{\textbf{#1}}}
\newcommand{\second}[1]{\textcolor{blue}{\underline{#1}}}

\title{RelayVSR: Large--Small Model Collaboration for Efficient Real-World Video Super-Resolution}

\author{\parbox[t]{\dimexpr\textwidth-2\tabcolsep\relax}{\centering
\textbf{Xijun Wang$^{1}$ \quad Xin Li$^{1}$ \quad
Zirui Lang$^{1}$ \quad Suhang Yao$^{1}$ \quad Haoran Li$^{1}$} \\
\textbf{Zhibo Chen$^{1}$} \\
\vspace{5pt}
{\normalfont $^{1}$University of Science and Technology of China}
}
}

\iclrfinalcopy
\begin{document}

\maketitle
\lhead{Preprint}

\begin{abstract}
Large generative models can recover realistic detail in real-world video super-resolution (VSR), but processing an entire video with them is computationally expensive.
In this work, we present \textbf{RelayVSR}, a streaming VSR framework built on the \textbf{Sparse Generative Relay} mechanism.
A large generative model generates reference latents for sparse keyframes, while a lightweight VSR network uses these references and low-resolution video to super-resolve every frame.
The lightweight VSR network, implemented as a Dual-Memory Video Transformer, reuses keyframe information across frames and updates recent video context, supporting first-keyframe conditioning and dual-endpoint conditioning with bounded lookahead.
However, errors in shared keyframes can propagate and accumulate across output frames, making keyframe quality alone an insufficient optimization target.
We address this collaboration gap with Video-Aware Reference Optimization (VARO), which uses reinforcement learning to update the large generative model with two reward levels: a system-level reward evaluates videos produced by the fixed lightweight VSR network, while a reference-level reward evaluates decoded keyframe quality.
VARO improves final video quality over direct joint training, and its dual-level rewards outperform a system-level reward alone.
At 1080p on a single NVIDIA A100 80GB, dual-endpoint RelayVSR with a 15-frame keyframe interval reaches 29.29 FPS, 13.82 GB peak GPU memory, and 0.327 s first-frame model latency, compared with 7.80 FPS, 24.447 GB, and 2.83 s for FlashVSR-Tiny.
The code is available at \url{https://github.com/kopperx/RelayVSR}.
 \end{abstract}

\section{Introduction}
\label{sec:introduction}

\vspace{-5pt}
Streaming video super-resolution (VSR) aims to produce high-quality, high-resolution video in real time from degraded low-resolution inputs~\citep{zhuang2026flashvsr}.
To improve output quality, recent diffusion-based VSR methods draw on pretrained video generation models.
Their learned priors help synthesize realistic textures and fine details~\citep{xie2025star}.
However, running these large models is computationally expensive, especially at high output resolutions~\citep{wang2025seedvr2}.

Recent work has reduced this cost by replacing iterative denoising with a single generative forward pass~\citep{chen2025dove,wang2025seedvr2}.
This avoids repeatedly evaluating a large model for the same video.
Further improvements target attention and decoding, two major sources of inference overhead.
Local or sparse attention reduces computation over video latents, while lightweight decoders accelerate high-resolution reconstruction~\citep{zhuang2026flashvsr,yan2026swiftvr}.
These advances have substantially reduced both sampling cost and the overhead of each forward pass.
However, even single-step inference in these approaches requires a large generative model to process spatiotemporal latents spanning the video sequence~\citep{zhuang2026flashvsr,yan2026swiftvr}.
Further optimizations of the generative backbone offer diminishing returns.
We therefore turn to a different question: \emph{Do we need a large generative model to process every frame for high-quality video super-resolution?}

\begin{figure}[t]
    \centering
    \includegraphics[width=\linewidth]{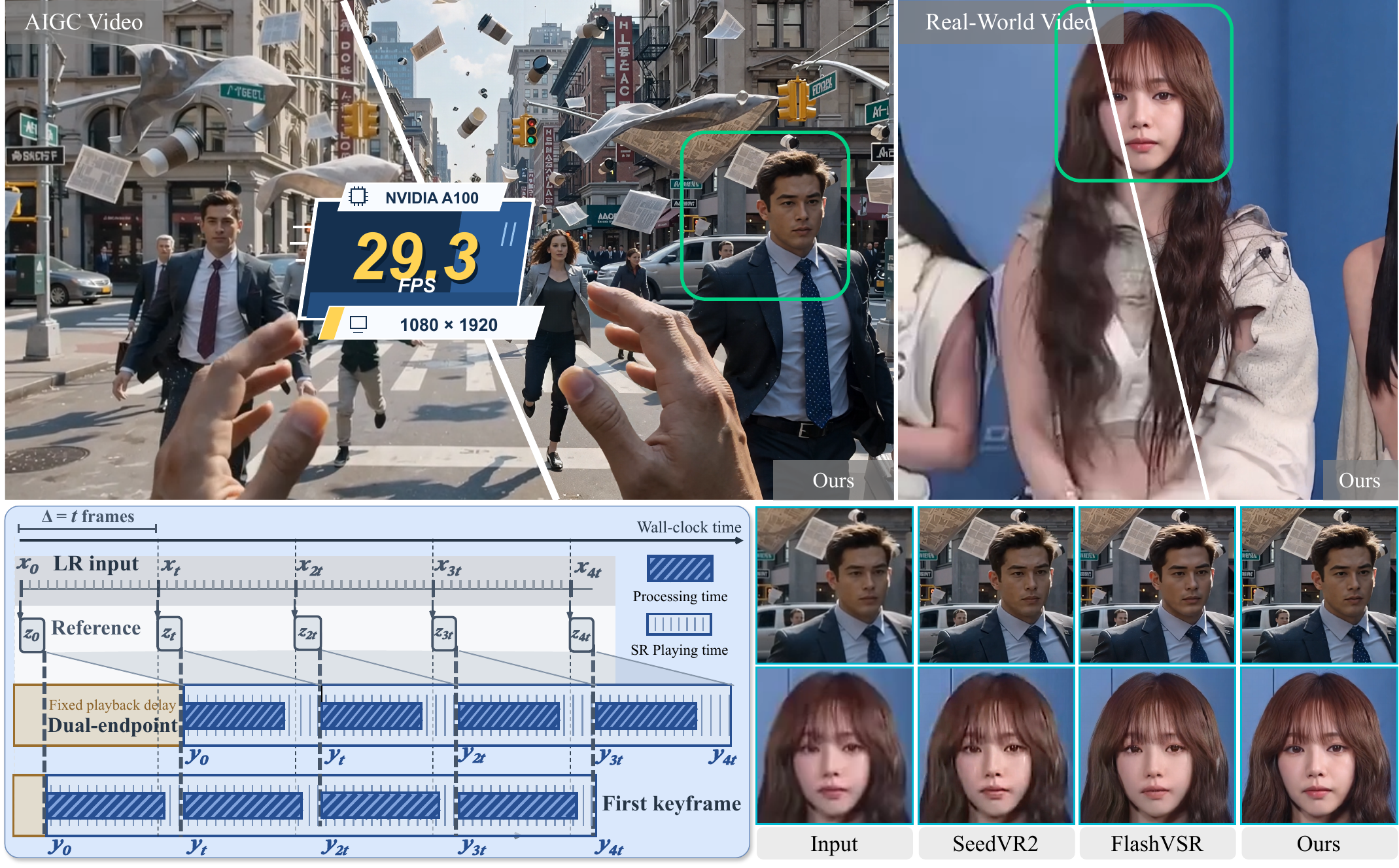}
    \setlength{\abovecaptionskip}{-6pt}
    \caption{\textbf{Visual comparison and streaming schedule of RelayVSR.}
    Top: RelayVSR outputs on an AI-generated video and a real-world video;
    the 29.3 FPS badge reports RelayVSR throughput at 1080p on one NVIDIA
    A100 80GB in dual-endpoint mode with a 15-frame keyframe interval.
    Bottom left: the sparse-reference schedule for first-keyframe and
    dual-endpoint inference, with a fixed playback delay in the latter.}
    \label{fig:intro}
    \vspace{-12pt}
\end{figure}

In this work, we introduce \textbf{RelayVSR}, an efficient streaming VSR framework built on a \textbf{(i) Sparse Generative Relay} mechanism.
A large generative model enhances only sparse keyframes, while a lightweight VSR network super-resolves every frame using the LR video and the generated reference latents.
The LR video provides frame-specific content and motion cues, while the enhanced keyframes supply fine details to guide super-resolution.
For each interval, RelayVSR can use the first keyframe alone to avoid input lookahead or both endpoints for richer reference guidance with bounded lookahead.
The keyframe interval determines how often the large generative model updates these references and therefore its amortized cost per output frame.
Fig.~\ref{fig:intro} illustrates this schedule alongside representative VSR results.
At 1080p on a single NVIDIA A100 80GB, the complete system in dual-endpoint mode with a 15-frame keyframe interval runs at 29.29 FPS with 13.82 GB peak GPU memory, compared with 7.80 FPS and 24.447 GB for FlashVSR-Tiny.

First, we adapt a pretrained video generation model~\citep{wan2025wan} to frame-wise target keyframe latents, followed by one-step GAN distillation with a relativistic adversarial objective~\citep{jolicoeur2019relativistic}.
Second, to exploit the rich generative priors encoded in the keyframe latents, we design a lightweight \textbf{(ii) Dual-Memory Video Transformer}.
The network super-resolves each video interval sequentially using keyframe latents from its first frame or both endpoints.
Persistent keyframe and rolling video memories cache keys and values from keyframes and recent frames, respectively. 
Current LR tokens query both memories to combine keyframe details with local temporal context.
Together with keyframe super-resolution, this design enables efficient streaming VSR while retaining the perceptual benefits of the large model’s generative priors.

However, sharing keyframe latents across frames also spreads the effects of keyframe errors.
Our experiments show that texture and color errors in super-resolved keyframes propagate to non-keyframes and accumulate during video super-resolution.
Adapting the lightweight VSR network to keyframe latents produced by the large model remains insufficient to prevent this propagation.
These findings expose an objective mismatch: optimizing keyframe quality alone does not account for how the lightweight network uses keyframe latents to super-resolve the full video.
We term this mismatch the \textbf{Collaboration Gap}.
Addressing it calls for optimizing keyframe super-resolution with feedback from the video produced by the lightweight network.

To address this gap, we introduce \textbf{(iii) Video-Aware Reference Optimization (VARO)}, which trains the large model using reward feedback from the video super-resolved by the lightweight VSR network.
During post-training, the system-level reward evaluates the HR videos super-resolved by the fixed lightweight network using the large model's keyframe latents, focusing on non-keyframe quality, temporal consistency, and continuity across keyframe updates.
The large model therefore learns from how its keyframe latents affect the frames that use them.
A complementary reference-level reward evaluates the decoded keyframes for content fidelity and realistic detail, maintaining a direct quality objective alongside downstream video feedback.
We implement VARO using DiffusionNFT~\citep{zheng2026diffusionnft}, updating only the large model with the combined reward.

Experiments demonstrate that RelayVSR achieves a favorable balance between super-resolution quality and inference efficiency compared with other diffusion-based VSR methods.
With reference keyframes spaced 15 frames apart, its dual-endpoint configuration processes 1080p video at 29.29 FPS on a single NVIDIA A100 80GB, with peak allocated GPU memory of 13.82 GB.
Ablations further show that VARO improves final video quality over direct joint training, and that combining reference-level and system-level rewards yields better results than system-level feedback alone.
 \vspace{-5pt}
\section{Related Work}
\label{sec:related_work}
\vspace{-3pt}
\subsection{Real-World Video Super-Resolution}
\vspace{-3pt}
Conventional VSR methods recover high-resolution frames through temporal alignment and feature aggregation, with real-world extensions addressing complex degradations~\citep{chan2022basicvsrpp,liang2022rvrt,chan2022realbasicvsr,zhang2024realviformer}.
Diffusion-based methods further exploit image and video generative priors for detail synthesis, with temporal modeling or motion guidance to maintain consistency~\citep{zhou2024upscaleavideo,yang2024mgldvsr,xie2025star,wang2025seedvr,li2025diffvsr}.
Their sampling cost has motivated one-step VSR methods that replace iterative denoising with a single forward pass~\citep{chen2025dove,wang2025seedvr2,sun2025dloral,li2025osdiffvsr}.
Beyond sampling, FlashVSR~\citep{zhuang2026flashvsr} reduces attention and decoding costs with locality-constrained sparse attention and an LR-conditioned lightweight decoder, while SwiftVR~\citep{yan2026swiftvr} combines shifted-window attention with a lightweight autoencoder for streaming inference.
Stream-DiffVSR~\citep{shiu2026streamdiffvsr} combines a distilled denoiser with motion-aligned temporal guidance for causal, frame-by-frame processing.
PS-SR~\citep{wu2026pssr} also distributes computation between models, using a large model for an initial denoising step and a lightweight model for subsequent refinements.
RelayVSR instead combines sparse keyframe super-resolution by a large generative model with lightweight frame-by-frame processing to enable fast, high-quality streaming VSR.
\vspace{-3pt}
\subsection{Reference-Guided Video Super-Resolution}
\vspace{-3pt}
Reference-guided VSR supplements LR videos with fine details from high-quality reference frames.
RefVSR and ERVSR obtain these references from an additional wide-angle camera to super-resolve ultra-wide videos~\citep{lee2022refvsr,kim2023ervsr}.
Within this multi-camera setting, RefVSR++~\citep{zou2025refvsrpp} improves temporal aggregation by separately propagating reference features and fused LR--reference features.
References can also be obtained from the LR video itself through image super-resolution.
DAM-VSR~\citep{kong2025damvsr} uses such references to guide appearance in video diffusion, while the LR sequence provides motion guidance.
SparkVSR~\citep{yu2026sparkvsr} similarly conditions a large video diffusion model on super-resolved keyframes, combining their sparse latents with LR video latents to enable interactive VSR.
In RelayVSR, generated keyframe latents instead guide a lightweight VSR network.
The network retains reference information in persistent keyframe memory and incorporates recent temporal context through rolling video memory, supporting frame-by-frame super-resolution.
\vspace{-3pt}
\subsection{Preference Optimization for Diffusion Models}
\vspace{-3pt}
Reinforcement learning and preference optimization enable diffusion models to learn from feedback beyond their original denoising objectives.
DDPO~\citep{black2024ddpo} formulates denoising as a sequential decision process and applies policy gradient optimization, while Diffusion-DPO~\citep{wallace2024diffusiondpo} learns directly from pairwise preferences.
DiffusionNFT~\citep{zheng2026diffusionnft} incorporates reward feedback into forward-process flow matching, enabling policy optimization without likelihood estimation.

For image super-resolution, RFSR~\citep{sun2024rfsr} combines perceptual reward feedback with structural constraints, while GDPO-SR~\citep{yi2026gdposr} adapts group-based preference optimization to one-step generative models.
Recent work further incorporates fidelity to the LR input: LucidNFT~\citep{fei2026lucidnft} combines perceptual and LR-referenced faithfulness rewards through DiffusionNFT, while OARS~\citep{zhao2026oars} evaluates fidelity preservation and perceptual gain in the LR-to-SR transition.
These methods optimize the quality of the SR images produced by the model being updated.
In RelayVSR, the large generative model produces reference latents for sparse keyframes to guide a lightweight VSR network.
The designed VARO optimizes the large generative model through reinforcement learning with two reward levels---a system-level reward and a reference-level reward---while keeping the lightweight VSR network fixed.
 \vspace{-6pt}
\section{Method}
\label{sec:method}

\begin{figure}[t]
    \centering
    \includegraphics[width=\linewidth]{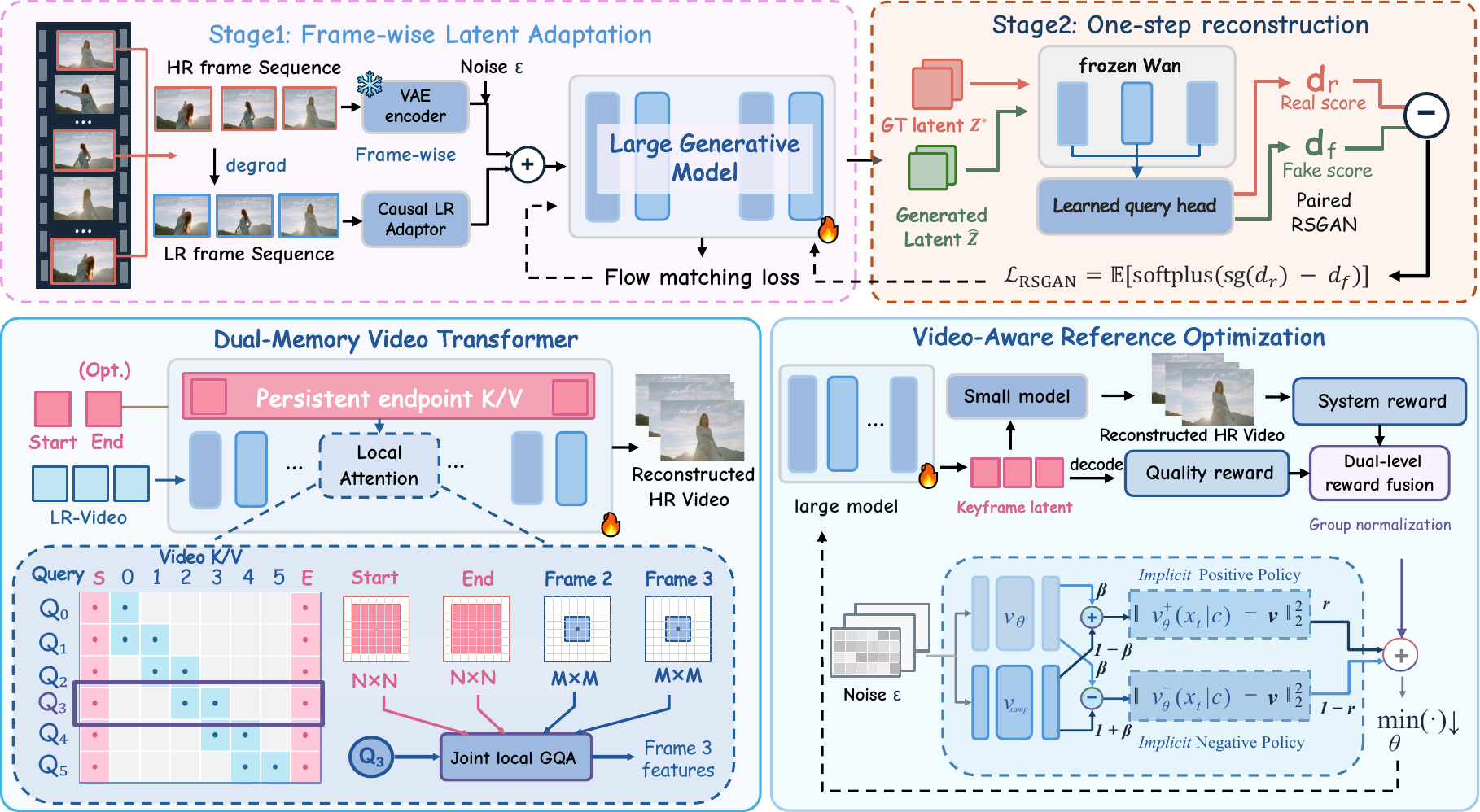}
    \setlength{\abovecaptionskip}{-6pt}
    \caption{\textbf{RelayVSR framework and training.} Top: the large generative model learns frame-wise target keyframe latents in Stage~1 and undergoes one-step GAN distillation in Stage~2. Bottom left: the lightweight VSR network super-resolves every frame by jointly attending to persistent keyframe and rolling video memories. Bottom right: VARO uses reinforcement learning to update the large generative model with a system-level reward on videos super-resolved by the fixed lightweight VSR network and a reference-level reward on decoded keyframes.}
    \label{fig:relayvsr}
    \vspace{-10pt}
\end{figure}
\vspace{-5pt}
\paragraph{Overview.}
RelayVSR assigns sparse keyframe super-resolution to a large generative model and uses the generated reference latents to guide a lightweight VSR network in super-resolving every frame (Fig.~\ref{fig:relayvsr}).
We first describe the Sparse Generative Relay (Sec.~\ref{sec:relay}) and the Dual-Memory Video Transformer that combines reusable references with recent video context (Sec.~\ref{sec:dual_memory}).
Finally, Video-Aware Reference Optimization (VARO) uses reinforcement learning to update the large generative model with a system-level reward and a reference-level reward, while keeping the lightweight VSR network fixed (Sec.~\ref{sec:varo}).

\vspace{-3pt}
\subsection{Sparse Generative Relay}
\label{sec:relay}
\vspace{-3pt}
Let $X=\{x_t\}_{t=1}^{T}$ denote the LR input video and $Y=\{y_t\}_{t=1}^{T}$ its HR training target.
We select keyframes every $\Delta$ frames starting at $k_1=1$ and denote their ordered indices by $\mathcal K=\{k_j\}_{j=1}^{J}$, where $\Delta$ is the keyframe interval.
The large generative model produces a reference latent $\hat z_j$ for each selected frame and passes it directly to the lightweight VSR network without RGB decoding.
The lightweight network uses these references and the LR video to super-resolve every frame, including those at keyframe positions.
\vspace{-3pt}
\paragraph{Keyframe super-resolution.}
We adapt the pretrained Wan2.2 video diffusion model~\citep{wan2025wan} with a learned LR projection through the two stages in Fig.~\ref{fig:relayvsr}.
In \emph{Stage 1 (frame-wise latent adaptation)}, we independently encode each HR keyframe as $z_j^\star=\mathcal E(y_{k_j})$ using the frozen VAE encoder $\mathcal E$.
We stack these targets into $Z^\star$ and fine-tune the model with LoRA using a flow-matching objective.
During training, a block-causal attention mask limits temporal attention to tokens from the current keyframe and recent keyframes within a fixed window.
In \emph{Stage 2}, we use GAN distillation to obtain a one-step reference generator, combining a latent reconstruction loss with a relativistic adversarial (RSGAN) objective~\citep{jolicoeur2019relativistic}.
Appendices~\ref{app:reference_generator} and~\ref{app:training_details} give the full objectives and training settings.
At inference, the generator $G_\theta$ uses key--value (KV) caching for streaming generation: $(\hat z_j,s_j^G)=G_\theta(x_{k_j},\epsilon_j,s_{j-1}^G)$, where $\epsilon_j$ is sampling noise and $s_j^G$ is the cache state.
\vspace{-3pt}
\paragraph{Streaming collaboration.}
Between consecutive keyframes $k_j$ and $k_{j+1}$, the lightweight VSR network super-resolves frames sequentially using the LR video and the selected reference latents.
In \emph{first-keyframe mode}, the network uses the left reference and causal LR context without future-frame dependency.
In \emph{dual-endpoint mode}, it also uses the right reference, introducing bounded input lookahead.
The large generative model runs only at keyframes, with $\Delta$ controlling its invocation frequency, while the lightweight network reuses the generated references to super-resolve every frame.
\vspace{-3pt}
\subsection{Dual-Memory Video Transformer}
\label{sec:dual_memory}
\vspace{-3pt}
Our lightweight VSR network, the Dual-Memory Video Transformer $F_\phi$, super-resolves each incoming LR frame using generated reference latents and information from recent frames.
We use a convolutional encoder to extract features from the LR frame and a separate projection to process the reference latents.
Both produce spatial tokens with the same feature dimension for the Transformer.
\vspace{-8pt}
\paragraph{Dual-memory attention.}
To reuse keyframe information across output frames, we process each reference independently, applying self-attention among its own spatial tokens at each Transformer layer.
Because this reference processing is independent of video features, we precompute its layer-wise keys and values and store them in a \emph{persistent keyframe memory}.
Super-resolving successive frames also requires recent temporal context to complement the reusable keyframe information.
We therefore maintain a \emph{rolling video memory} that stores the layer-wise keys and values of recently processed video frames and is updated after each frame.

At each Transformer layer, the current frame jointly attends to the selected references, recent video history, and its own features.
Let $Q_t$ denote the current-frame queries, $(K_t^r,V_t^r)$ the key and value projections of the selected references, and $(K_t^v,V_t^v)$ those of the recent video history and the current frame.
The attention output is
\begin{equation}
    a_t=\operatorname{Attn}\!\left(Q_t,[K_t^r;K_t^v],[V_t^r;V_t^v]\right),
    \label{eq:dual_memory_attention}
\end{equation}
where $[\,;\,]$ concatenates tokens.
We restrict attention to local spatial neighborhoods to limit computation.
Since sparse keyframes can be farther away in time than recent video frames, we use a larger neighborhood when reading references to accommodate potentially greater spatial displacement.
\vspace{-3pt}
\paragraph{Super-resolution and training.}
The output head uses a projection and PixelShuffle to convert the final features into an RGB residual at the target resolution.
Adding this residual to the bicubic-upsampled LR frame gives the $4\times$ super-resolved output.
We train the network from scratch with reference latents encoded from ground-truth HR keyframes, using $L_1$ and LPIPS losses~\citep{zhang2018lpips} on all output frames.
During training, we randomly omit the right reference so that the same network supports first-keyframe and dual-endpoint conditioning.
Appendix~\ref{app:small_network} details the attention implementation, and Appendix~\ref{app:base_training_config} lists the supervised training settings.
\vspace{-3pt}
\subsection{Video-Aware Reference Optimization}
\label{sec:varo}
\vspace{-3pt}
The lightweight VSR network is trained with reference latents encoded from ground-truth HR keyframes, whereas inference uses references generated from LR keyframes by the large generative model.
Texture and color errors in these generated references can propagate and accumulate as they are reused across frames.
Keyframe-level objectives do not capture these downstream effects.
VARO addresses this Collaboration Gap through reinforcement learning with system-level and reference-level rewards, optimizing the large generative model for both reference and video quality while keeping the lightweight VSR network fixed.
\vspace{-3pt}
\paragraph{Dual-level rewards.}
For each LR--HR training clip $(X,Y)$, the one-step generator samples a causal sequence of reference latents $\hat Z$.
Using the streaming procedure in Sec.~\ref{sec:relay}, the lightweight VSR network with fixed parameters $\phi^\star$ produces $\hat Y=F_{\phi^\star}(X,\hat Z)$.
The \emph{system-level reward} $R_{\mathrm{sys}}$ scores the entire super-resolved video, including keyframe outputs.
It combines RGB $L_1$ fidelity across all frames, DOVER++ technical video quality~\citep{wu2023dover}, MUSIQ~\citep{ke2021musiq} on temporally spaced frames, and optical-flow warping error over consecutive output frames.
The \emph{reference-level reward} $R_{\mathrm{ref}}$ evaluates the same generated latents after decoding by the frozen VAE decoder $\mathcal D$.
It combines RGB $L_1$ against each HR keyframe $y_{k_j}$, MUSIQ on every decoded reference $\mathcal D(\hat z_j)$, and optical-flow warping error between successive decoded references.

Each reference rollout receives the combined reward
\begin{equation}
    R_{\mathrm{total}}=\lambda_{\mathrm{sys}}\widetilde R_{\mathrm{sys}}(\hat Y,Y)+\lambda_{\mathrm{ref}}\widetilde R_{\mathrm{ref}}(\hat Z,Y_{\mathcal K}),
    \label{eq:varo_objective}
\end{equation}
where $Y_{\mathcal K}$ denotes the HR keyframes, the tildes denote reward-scale normalization, and the weights balance video and reference quality.
\vspace{-3pt}
\paragraph{Generator optimization.}
We aim to maximize $\mathbb E_{(X,Y),\,\hat Z\sim p_\theta(\cdot\mid X_{\mathcal K})}[R_{\mathrm{total}}]$ using DiffusionNFT~\citep{zheng2026diffusionnft}.
For a sampled reference sequence $u=\hat Z$, its group-normalized reward is mapped to $r\in[0,1]$.
At the fixed noise endpoint $\tau=1$, fresh noise $\epsilon\sim\mathcal N(0,I)$ gives the velocity target $v=\epsilon-u$.
We minimize
\begin{equation}
    \mathcal L_{\mathrm{VARO}}=\mathbb E\!\left[r\|v_\theta^+-v\|_2^2+(1-r)\|v_\theta^--v\|_2^2\right],
    \label{eq:varo_nft}
\end{equation}
where $v_\theta^\pm=v_{\mathrm{samp}}\pm\beta(v_\theta-v_{\mathrm{samp}})$, with fixed sampling-policy prediction $v_{\mathrm{samp}}$, trainable prediction $v_\theta$, and mixing coefficient $\beta$.
Appendix~\ref{app:varo_details} gives the reward formulas, and Appendix~\ref{app:varo_settings} lists the training settings; inference retains one-step reference generation and streaming VSR.
 \vspace{-3pt}
\section{Experiments}
\label{sec:experiments}
\vspace{-3pt}
\subsection{Experimental Setup}
\vspace{-3pt}
\paragraph{Implementation Details.}
RelayVSR uses Wan2.2-TI2V-5B as its large generative model and a 31.9M-parameter lightweight VSR network; training draws on approximately 0.5M videos and 1M images.
We adapt the large model with rank-512 LoRA for 75K flow-matching steps, followed by 5K steps of one-step GAN distillation.
Both stages use eight-frame video clips sampled with random temporal strides and single images, with $704\times1280$ HR crops and batch sizes of 32 and 96, respectively.
We use AdamW with learning rates of 2e-5 for flow matching and 1e-5 for GAN distillation.
The lightweight VSR network is trained from scratch for 60K steps on 16-frame clips with $1024\times1536$ HR crops and a batch size of 64, using AdamW with a learning rate of 1e-4.
VARO then updates the large model with DiffusionNFT while keeping the lightweight VSR network fixed.
Additional training details are provided in Appendix~\ref{app:training_details}.
\vspace{-3pt}
\paragraph{Evaluation Protocol.}
We assess $4\times$ VSR on three synthetic benchmarks (REDS30, UDM10, and YouHQ40) and the real-world VideoLQ benchmark.
For the synthetic sets, LR inputs are generated from HR videos with the same RealBasicVSR degradation pipeline~\citep{chan2022realbasicvsr} used for training.
For sustained streaming, we use LongVSR60, which contains 30 real-world and 30 AI-generated single-shot videos of approximately 1,000 frames each.
The main comparison includes RealViformer~\citep{zhang2024realviformer}, STAR~\citep{xie2025star}, DOVE~\citep{chen2025dove}, SeedVR2~\citep{wang2025seedvr2}, SparkVSR~\citep{yu2026sparkvsr}, SwiftVR~\citep{yan2026swiftvr}, and FlashVSR~\citep{zhuang2026flashvsr}.
Throughout the paper, FlashVSR refers to its Tiny v1.1 variant.
For paired synthetic videos, we measure reconstruction fidelity with PSNR and SSIM and perceptual similarity with LPIPS.
We assess no-reference quality with NIQE, MUSIQ, CLIP-IQA, and DOVER on both synthetic and real-world videos.
A blind pairwise user study is described in Appendix~\ref{app:user_study}.
\vspace{-3pt}
\subsection{Comparison with Existing Methods}
\label{sec:comparison}
\vspace{-3pt}
\begin{table}[!b]
\begingroup
\centering
\caption{Quantitative comparison on synthetic and real-world VSR benchmarks.
The best and second-best results are marked in \best{red} and \second{blue}, respectively.}
\label{tab:main_results}
\renewcommand{\arraystretch}{1.1}  \resizebox{\textwidth}{!}{
\setlength{\heavyrulewidth}{1.2pt}
\begin{tabular}{cl|ccccccc|c}
\toprule
Dataset & Metric & RealViformer & STAR & DOVE & SeedVR2 & SparkVSR & SwiftVR & FlashVSR & Ours \\
\midrule

\multirow{7}{*}{\textbf{REDS30}}
 & PSNR $\uparrow$ & \best{23.32} & 22.14 & 23.28 & 22.17 & \second{23.29} & 21.33 & 21.41 & 22.77 \\
 & SSIM $\uparrow$ & 0.5967 & 0.5432 & \best{0.6103} & 0.5860 & \second{0.6050} & 0.5234 & 0.5399 & 0.5874 \\
 & LPIPS $\downarrow$ & \best{0.3043} & 0.4967 & 0.3773 & 0.3158 & 0.3909 & 0.3564 & 0.3311 & \second{0.3051} \\
 & NIQE $\downarrow$ & \second{3.0804} & 5.2237 & 4.2282 & 3.5157 & 4.4496 & 3.4153 & \best{2.9504} & 3.5398 \\
 & MUSIQ $\uparrow$ & \second{59.12} & 37.95 & 50.44 & 57.65 & 47.58 & \best{63.77} & 56.01 & 54.30 \\
 & CLIP-IQA $\uparrow$ & \second{0.3236} & 0.2132 & 0.2904 & 0.3041 & 0.2755 & \best{0.3914} & 0.3160 & 0.3161 \\
 & DOVER $\uparrow$ & 0.3293 & 0.2068 & 0.3337 & 0.3679 & 0.3200 & \second{0.3893} & 0.3477 & \best{0.4010} \\
\midrule

\multirow{7}{*}{\textbf{UDM10}}
 & PSNR $\uparrow$ & \best{26.65} & 25.49 & \second{26.39} & 25.99 & 26.31 & 25.41 & 24.41 & 25.94 \\
 & SSIM $\uparrow$ & 0.7483 & 0.7237 & \second{0.7504} & 0.7313 & 0.7430 & 0.7213 & 0.7008 & \best{0.7545} \\
 & LPIPS $\downarrow$ & 0.2670 & 0.3677 & 0.2341 & \best{0.2150} & 0.2554 & 0.2640 & 0.2528 & \second{0.2171} \\
 & NIQE $\downarrow$ & 4.4199 & 7.1730 & 4.7713 & 4.6516 & 4.7099 & \second{4.2548} & \best{4.0032} & 4.3205 \\
 & MUSIQ $\uparrow$ & 56.98 & 31.50 & 60.30 & 58.49 & 58.28 & \second{64.55} & 64.24 & \best{65.32} \\
 & CLIP-IQA $\uparrow$ & 0.3705 & 0.2225 & 0.4717 & 0.4000 & 0.4687 & \second{0.4986} & \best{0.4987} & 0.4852 \\
 & DOVER $\uparrow$ & 0.4646 & 0.2455 & 0.4996 & 0.5013 & 0.4823 & 0.4744 & \best{0.5507} & \second{0.5448} \\
\midrule

\multirow{7}{*}{\textbf{YouHQ40}}
 & PSNR $\uparrow$ & 24.17 & 23.46 & \best{24.29} & 23.66 & \second{24.21} & 23.08 & 22.68 & 23.19 \\
 & SSIM $\uparrow$ & 0.6414 & 0.6429 & \best{0.6662} & 0.6555 & \second{0.6605} & 0.6151 & 0.6026 & 0.6340 \\
 & LPIPS $\downarrow$ & 0.3408 & 0.4545 & 0.2920 & \second{0.2726} & 0.3152 & 0.2912 & 0.2742 & \best{0.2569} \\
 & NIQE $\downarrow$ & 3.6506 & 6.9886 & 4.3244 & 4.1732 & 4.5953 & \second{3.2815} & \best{3.1768} & 3.5769 \\
 & MUSIQ $\uparrow$ & 59.63 & 32.91 & 60.63 & 57.69 & 56.67 & 62.01 & \best{65.79} & \second{63.21} \\
 & CLIP-IQA $\uparrow$ & 0.3996 & 0.2651 & 0.4500 & 0.3970 & 0.4051 & 0.5128 & \best{0.5333} & \second{0.5182} \\
 & DOVER $\uparrow$ & 0.6149 & 0.4308 & 0.6649 & 0.6709 & 0.6287 & 0.6877 & \second{0.6916} & \best{0.7313} \\
\midrule

\multirow{4}{*}{\textbf{VideoLQ}}
 & NIQE $\downarrow$ & \second{4.3616} & 5.6624 & 5.0218 & 4.7326 & 5.0373 & 4.4435 & \best{3.9488} & 4.4729 \\
 & MUSIQ $\uparrow$ & 49.22 & 35.29 & 44.91 & 41.68 & 42.48 & 50.46 & \second{50.73} & \best{51.26} \\
 & CLIP-IQA $\uparrow$ & 0.3263 & 0.2454 & 0.2942 & 0.2428 & 0.2890 & \second{0.3583} & \best{0.3623} & 0.3458 \\
 & DOVER $\uparrow$ & 0.4621 & 0.4135 & 0.4972 & 0.4455 & 0.4797 & 0.4957 & \second{0.5339} & \best{0.5488} \\

\bottomrule
\end{tabular}
}
\par
\endgroup
\vspace{-6pt}
\end{table}
 
\begin{figure}[t]
    \centering
    \includegraphics[width=\linewidth]{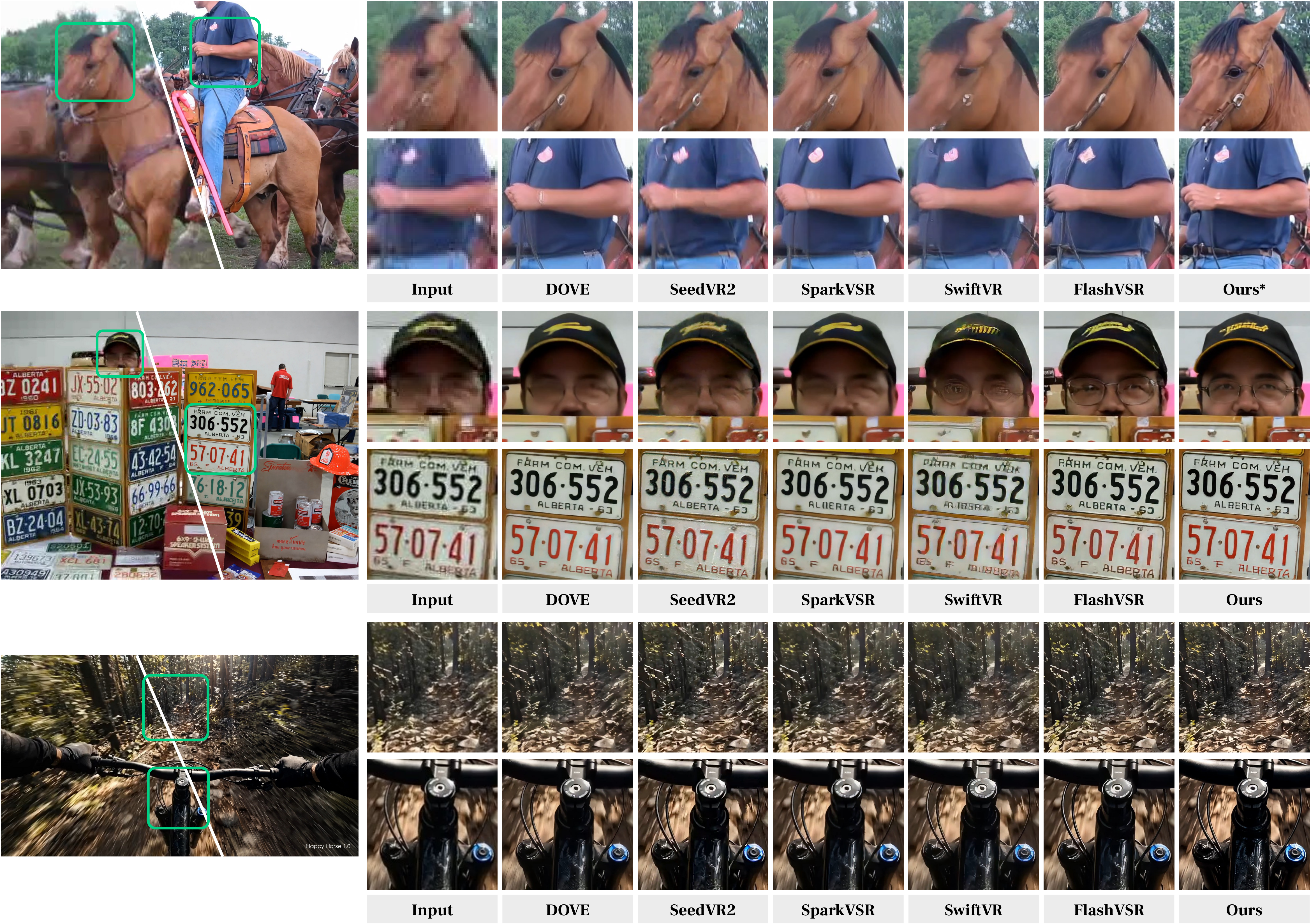}
    \setlength{\abovecaptionskip}{-5pt}
    \caption{\textbf{Qualitative VSR comparisons.} The top two examples are real-world videos and the bottom example is an AIGC video. $*$ marks the keyframe sample.}
    \label{fig:qualitative}
    \vspace{-8pt}
\end{figure}

\begin{figure}[t]
    \centering
    \includegraphics[width=\linewidth]{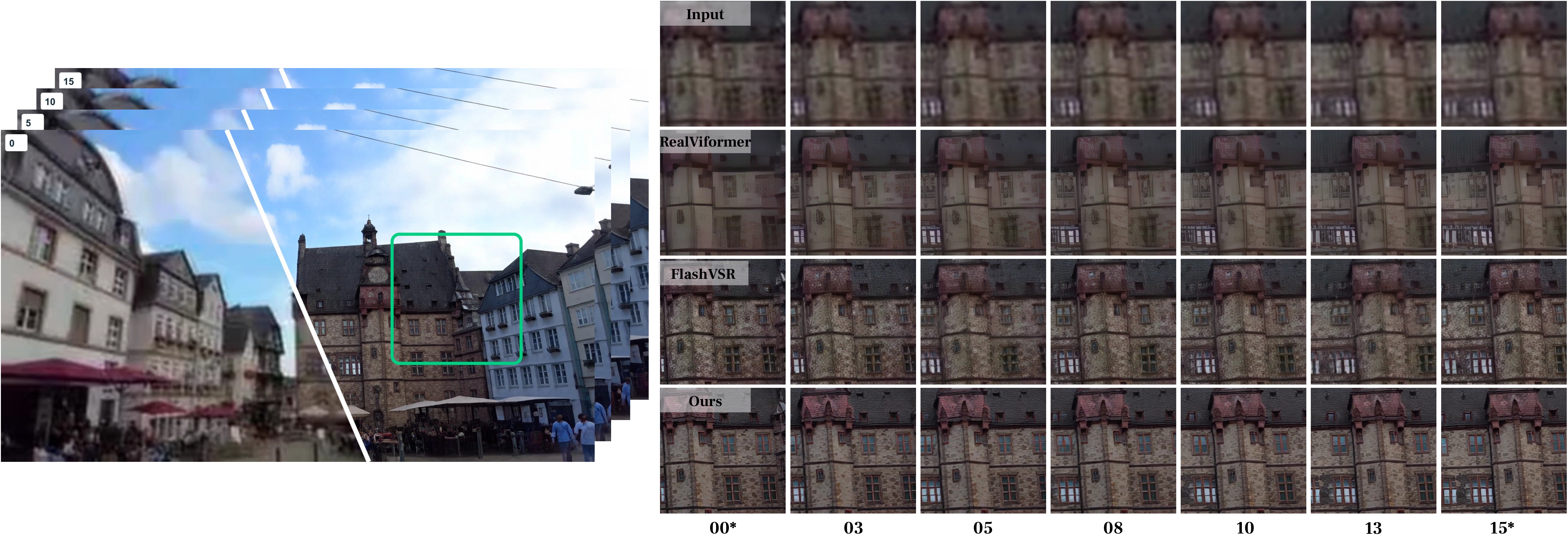}
    \setlength{\abovecaptionskip}{-5pt}
    \caption{\textbf{Keyframe and non-keyframe comparison on a real-world video.} The $*$ symbol marks keyframes; the other columns are non-keyframes.}
    \label{fig:keyframe_compare}
    \vspace{-12pt}
\end{figure}

\paragraph{Quantitative Comparisons.}
Table~\ref{tab:main_results} shows that RelayVSR maintains strong full-video quality with dual-endpoint conditioning and large-model calls spaced 15 frames apart.
Across the three paired benchmarks, it surpasses FlashVSR in PSNR, SSIM, and LPIPS.
It achieves the best LPIPS (0.2569) and DOVER (0.7313) on YouHQ40, the best DOVER on REDS30 (0.4010), and the best SSIM (0.7545) and MUSIQ (65.32) on UDM10.
On real-world VideoLQ, RelayVSR leads in MUSIQ (51.26) and DOVER (0.5488), while FlashVSR performs better in NIQE and CLIP-IQA.
The evaluation covers complete output videos rather than keyframes alone.
\vspace{-3pt}
\paragraph{Qualitative Comparisons.}
Fig.~\ref{fig:qualitative} compares RelayVSR with five generative VSR methods on two real-world videos and one AIGC video.
RelayVSR delineates the horse's face and bridle, retains recognizable facial features and license-plate characters in the second example, and resolves bicycle components and foliage in the AIGC cycling scene.
To examine whether detail persists between keyframes, Fig.~\ref{fig:keyframe_compare} follows a building crop between keyframes 00 and 15 through five intermediate frames.
Window outlines, roof edges, and masonry remain clear at these non-keyframe positions, illustrating that sparse generative references support visual quality throughout the interval.

\vspace{-3pt}
\subsection{System Efficiency Analysis}
\label{sec:efficiency_streaming}
\vspace{-2pt}
\begin{wraptable}{r}{0.5\textwidth}
    \vspace{-20pt}
    \centering
    \caption{Inference efficiency at $1080\times1920$ output resolution on one NVIDIA A100-80G using 201-frame clips. Timing excludes input wait and operations outside model inference.}
    \label{tab:system_efficiency}
    \small
    \setlength{\tabcolsep}{4pt}
    \resizebox{\linewidth}{!}{\begin{tabular}{@{}lccc@{}}
    \toprule
    \multirow{2}{*}{Method} & \multirow{2}{*}{FPS $\uparrow$} & Peak Mem. & First output \\
    & & (GB) $\downarrow$ & (s) \\
    \midrule
    DOVE & 0.56 & 41.08 & 356.801 \\
    SeedVR2-3B & 0.79 & 73.89 & 255.199 \\
    Stream-DiffVSR & 0.85 & 25.92 & 0.705 \\
    SwiftVR & 23.48 & 31.36 & 1.107 \\
    FlashVSR & 7.80 & 24.45 & 2.830 \\
    \midrule
    Ours (first-keyframe) & \textbf{35.89} & \textbf{13.72} & \textbf{0.183} \\
    Ours (dual-endpoint) & \underline{29.29} & \underline{13.82} & \underline{0.327} \\
    \bottomrule
    \end{tabular}}
    \vspace{-6pt}
\end{wraptable}
\paragraph{Inference efficiency.}
Table~\ref{tab:system_efficiency} shows the efficiency gains from invoking the large generative model only at sparse keyframes.
On 201-frame clips at 1080p using one NVIDIA A100-80G, RelayVSR in dual-endpoint mode reaches 29.29 FPS with 13.82\,GB peak allocated memory.
This is $3.76\times$ the throughput of FlashVSR (7.80 FPS, 24.45\,GB) while using 43.5\% less memory; SwiftVR, the fastest baseline in the table, reaches 23.48 FPS.
Switching to first-keyframe mode raises throughput to 35.89 FPS and reduces peak memory to 13.72\,GB; the first output requires 0.183\,s of model computation, compared with 0.327\,s in dual-endpoint mode.
In a real streaming system, dual-endpoint mode must also wait for the future keyframe to arrive---up to 0.500\,s at a 30-FPS input rate---before it can output frames in that interval.

\begin{wrapfigure}[13]{R}{0.44\textwidth}
    \vspace{-4.5pt}
    \centering
    \includegraphics[width=\linewidth]{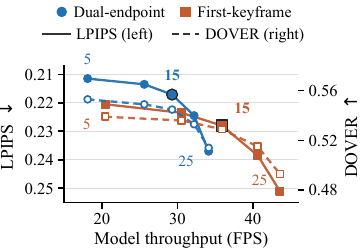}
    \vspace{-20pt}
    \caption{\textbf{Quality--efficiency trade-off.} Left to right: $\Delta=5,10,15,20,25$.}
    \label{fig:quality_efficiency}
    \vspace{-10pt}
\end{wrapfigure}
\paragraph{Quality--efficiency trade-off.}
Fig.~\ref{fig:quality_efficiency} shows diminishing returns as the keyframe interval increases.
In dual-endpoint mode, increasing $\Delta$ from 5 to 15 raises 1080p throughput from 18.10 to 29.29 FPS while UDM10 LPIPS rises from 0.2115 to 0.2171 and DOVER falls from 0.5530 to 0.5448.
Increasing it to 25 adds only 4.84 FPS, but LPIPS reaches 0.2370 and DOVER drops to 0.5140.
First-keyframe mode is faster at each interval with higher LPIPS and lower DOVER; at $\Delta=15$, it reaches 35.89 FPS, 0.2280 LPIPS, and 0.5290 DOVER.
These results support dual-endpoint $\Delta=15$ for the main comparisons.
\mbox{Appendix~\ref{app:relay_inference_efficiency}} reports timings across resolutions.

\pagebreak[4]
\vspace{-3pt}
\subsection{Ablation Study}
\label{sec:ablation}
\vspace{-3pt}
The efficiency gains above rely on each sparse reference guiding many lightweight VSR updates.
We therefore examine how reference and video memories contribute to video quality, then test whether VARO makes the shared references more useful to the final video (Tables~\ref{tab:memory_ablation} and~\ref{tab:varo_ablation}).

\begin{wraptable}{r}{0.55\textwidth}
\vspace{-8pt}
\begingroup
\centering
\caption{\textbf{Memory ablation} on UDM10 before VARO ($\Delta=15$, dual-endpoint). \ding{51}/\ding{55} indicates enabled/disabled memory; MS is motion smoothness.}
\label{tab:memory_ablation}
\small
\setlength{\tabcolsep}{2pt}
\renewcommand{\arraystretch}{1.12}
\begin{tabular*}{\linewidth}{@{\extracolsep{\fill}}ccccc@{}}
\toprule
Ref. memory & Video memory & MUSIQ $\uparrow$ & DOVER $\uparrow$ & MS $\uparrow$ \\
\midrule
\ding{55} & \ding{51} & 56.80 & 0.4760 & \textbf{0.9911} \\
\ding{51} & \ding{55} & \underline{63.65} & \underline{0.5110} & 0.9769 \\
\ding{51} & \ding{51} & \textbf{63.72} & \textbf{0.5290} & \underline{0.9830} \\
\bottomrule
\end{tabular*}
\endgroup
\vspace{-5pt}
\end{wraptable}

\paragraph{Reference and video memory.}
The two memory streams play complementary roles in sparse-reference VSR.
Table~\ref{tab:memory_ablation} removes either reference memory or rolling video memory from the lightweight VSR network.
All variants use the same training data, budget, and inference schedule; variants with references share the same pre-VARO one-step generator.
Adding references to the video-memory-only variant raises MUSIQ from 56.80 to 63.72 and DOVER from 0.4760 to 0.5290, showing the contribution of generative keyframe information to output quality.
Adding video memory to the reference-only variant further improves DOVER from 0.5110 to 0.5290 and MS from 0.9769 to 0.9830, while retaining similar MUSIQ (63.65 vs.\ 63.72).
Video memory alone has the highest MS (0.9911), while dual memory yields the best MUSIQ and DOVER and improves MS over reference memory alone.

\begin{wraptable}{r}{0.55\textwidth}
\vspace{-8pt}
\begingroup
\centering
\caption{\textbf{VARO ablation} on UDM10 ($\Delta=15$, dual-endpoint). MS is motion smoothness.}
\label{tab:varo_ablation}
\small
\setlength{\tabcolsep}{2pt}
\renewcommand{\arraystretch}{1.12}
\begin{tabular*}{\linewidth}{@{\extracolsep{\fill}}lcccc@{}}
\toprule
\multirow[c]{2}{*}{Variant} & \multicolumn{2}{c}{MUSIQ $\uparrow$} & \multirow[c]{2}{*}{DOVER $\uparrow$} & \multirow[c]{2}{*}{MS $\uparrow$} \\
\cmidrule(lr){2-3}
 & All & Key & & \\
\midrule
w/o post-training & 63.72 & 64.05 & 0.5290 & 0.9830 \\
\addlinespace[3pt]
\begin{tabular}[c]{@{}l@{}}w/ reference-level\\reward only\end{tabular} & 64.20 & \underline{64.95} & 0.5355 & 0.9838 \\
\addlinespace[3pt]
\begin{tabular}[c]{@{}l@{}}w/ system-level\\reward only\end{tabular} & \underline{64.86} & 64.42 & \underline{0.5400} & \underline{0.9851} \\
\addlinespace[3pt]
\textbf{Ours} & \textbf{65.32} & \textbf{65.18} & \textbf{0.5448} & \textbf{0.9900} \\
\bottomrule
\end{tabular*}
\endgroup
\vspace{-5pt}
\end{wraptable}

\paragraph{Video-aware reference optimization.}
VARO improves references according to the video they help produce.
Table~\ref{tab:varo_ablation} compares reward variants initialized from the same one-step GAN generator, with the lightweight VSR network fixed.
Full VARO raises all-frame MUSIQ from 63.72 to 65.32, DOVER from 0.5290 to 0.5448, and MS from 0.9830 to 0.9900 over the model without post-training.
The two reward levels favor different aspects: reference-level feedback alone yields higher keyframe MUSIQ than system-level feedback (64.95 vs.\ 64.42), whereas system-level feedback gives stronger all-frame MUSIQ, DOVER, and MS.
Combining them improves all four reported scores, including keyframe MUSIQ to 65.18.
Thus, evaluating both the generated references and the final video provides a more effective training signal for large--small model collaboration.
 \vspace{-3pt}
\section{Conclusion}
\label{sec:conclusion}
RelayVSR makes generative real-world video super-resolution efficient by sharing sparse large-model outputs across frames.
A large generative model supplies keyframe reference latents, and a lightweight VSR network uses dual memory to carry their detail through the video while tracking recent temporal context.
Video-Aware Reference Optimization (VARO) trains the large generative model using feedback on both decoded keyframes and the resulting video, making the references more useful to the lightweight VSR network.
Experiments show that this collaboration delivers strong full-video quality while amortizing large-model computation across frames.
RelayVSR leads the compared methods in LPIPS and DOVER on YouHQ40 and in MUSIQ and DOVER on real-world VideoLQ.
At 1080p with $\Delta=15$, dual-endpoint mode reaches 29.29 model-computation FPS with 13.82\,GB peak allocated GPU memory; first-keyframe mode reaches 35.89 FPS without future-frame input.
These results demonstrate how reusable references and downstream optimization extend the benefits of a large generative model across an entire video.
Adapting reference placement to scene dynamics is a promising direction for further improving the quality--latency trade-off.

\appendix
\clearpage

\begin{center}
    \Large\textbf{{Appendix}}\\
    \vspace{8mm}
\end{center}

\section{Additional Method Details}
\label{app:method_details}

\subsection{Reference Generation}
\label{app:reference_generator}

\paragraph{Flow-matching objective.}
For the independently encoded targets $Z^\star$ defined in Sec.~\ref{sec:relay}, normalized diffusion time $\tau\in[0,1]$, and Gaussian noise $\epsilon\sim\mathcal N(0,I)$, we form
\begin{equation}
    Z_\tau=(1-\tau)Z^\star+\tau\epsilon,
    \qquad v^\star=\epsilon-Z^\star,
\end{equation}
then optimize the velocity prediction with
\begin{equation}
    \mathcal L_{\mathrm{FM}}
    =\mathbb E_{Z^\star,c,\tau,\epsilon}
    \big[\operatorname{MSE}(v_\theta(Z_\tau,\tau,c),v^\star)\big],
    \label{eq:app_flow_matching}
\end{equation}
where $c$ contains projected LR keyframes and their available temporal context.
The block-causal attention mask restricts each keyframe to a window of $H$ keyframes, including itself.

\paragraph{One-step adversarial objective.}
We use a relativistic adversarial objective~\citep{jolicoeur2019relativistic} in addition to latent reconstruction.
Let $d_{\mathrm{real}}$ and $d_{\mathrm{gen}}$ be the discriminator logits and $\sigma$ the sigmoid function.
The two adversarial losses are
\begin{equation}
\begin{aligned}
    \mathcal L_{\mathrm{RSGAN}}^G
    &=-\mathbb E\log\sigma(d_{\mathrm{gen}}-d_{\mathrm{real}}),\\
    \mathcal L_{\mathrm{RSGAN}}^D
    &=-\mathbb E\log\sigma(d_{\mathrm{real}}-d_{\mathrm{gen}}).
\end{aligned}
\end{equation}
The generator and discriminator objectives are
\begin{equation}
\begin{aligned}
    \mathcal L_G^{\mathrm{1step}}
    &=\operatorname{MSE}(\hat Z,Z^\star)
      +\lambda_{\mathrm{adv}}\mathcal L_{\mathrm{RSGAN}}^G,\\
    \mathcal L_D
    &=\mathcal L_{\mathrm{RSGAN}}^D
      +\gamma_{\mathrm{R1}}\mathcal L_{\mathrm{feature\text{-}R1}},
\end{aligned}
    \label{eq:keyframe_training}
\end{equation}
The discriminator scores intermediate features of a frozen Wan feature extractor with a trainable pooling head; feature-space R1 regularizes gradients with respect to those input features.

During streaming inference, the generator attends to a local spatial neighborhood around each keyframe latent token and uses a rolling KV cache for recent keyframes; rotary position encoding (RoPE) is applied to cached keys when they are read.

\subsection{Dual-Memory Attention Implementation}
\label{app:small_network}

\paragraph{Token resolution.}
For a spatially padded LR input of size $h\times w$, the LR encoder reduces the grid to $h/4\times w/4$.
The frozen VAE encodes a $4\times$ HR keyframe with spatial stride 16, giving a reference latent on the same grid.
The LR and reference streams use separate input RMS normalization and type embeddings before entering the shared Transformer width.

\paragraph{Reference feature reuse.}
Reference tokens use spatial RoPE and a common reference type embedding, with no left/right endpoint role encoding.
Their layer-wise K/V therefore depend on the reference content rather than its role in an interval; the cached right endpoint can serve as the next interval's left endpoint without being encoded again.

\paragraph{Joint attention rules.}
The eligible reference, history, and current-frame K/V in Eq.~(\ref{eq:dual_memory_attention}) participate in one grouped-query attention softmax~\citep{ainslie2023gqa}.
Reference and video streams have separate K/V projections and RMS-normalized K/V features, while sharing the attention normalization, query and output projections, and SwiGLU feed-forward network.
Recent video frames are close in time, so a small neighborhood focuses on nearby motion cues at low attention cost.
Sparse references may be $\Delta$ frames away; a wider neighborhood lets each query search for corresponding details across larger displacements.
Both neighborhoods are centered on the current query, keeping the search local rather than attending to the entire frame; at image boundaries, only valid tokens are included.
The current video query has temporal RoPE position zero, while retained video frames have negative relative offsets; reference tokens use only spatial RoPE.
At most $W-1$ preceding video frames can contribute K/V to a query, with $W$ including the current frame.

\subsection{VARO Reward Function}
\label{app:varo_details}

Let $q_{\ell,m}$ be a calibrated component score, oriented so that higher is better, and let $w_{\ell,m}\geq0$ be its weight, with $\ell=s$ for the system and $\ell=r$ for references.
The system and reference rewards are
\begin{equation}
    R_{\mathrm{sys}}=\sum_{m\in\{L,D,M,W\}}w_{s,m}q_{s,m},\qquad
    R_{\mathrm{ref}}=\sum_{m\in\{L,M,W\}}w_{r,m}q_{r,m}.
    \label{eq:varo_rewards}
\end{equation}
Here $L$ is negative RGB $L_1$ error, $D$ is the technical-branch DOVER++ score~\citep{wu2023dover}, $M$ is MUSIQ~\citep{ke2021musiq}, and $W$ is negative flow-guided warping error.
System scores use the emitted video; reference scores use decoded keyframes.
The warping term compares consecutive emitted frames, including reference updates, or successive decoded keyframes, respectively.
The two level scores enter the combined objective in Eq.~(\ref{eq:varo_objective}).
 
\section{Training Details}
\label{app:training_details}

\subsection{Model Configuration}
\label{app:network_config}

During base training, the Wan backbone remains frozen: the large generative model updates its LoRA adapters and LR projection, and the GAN discriminator learns a pooling head over frozen Wan features.
The lightweight VSR network is trained end to end.
Table~\ref{tab:app_network_configuration} reports the dimensions and local attention windows of these components.

\begin{table}[!h]
\centering
\caption{\textbf{Model configuration.} Parameter counts include trainable parameters only; spatial attention windows are measured in tokens.}
\label{tab:app_network_configuration}
\small
\setlength{\tabcolsep}{5pt}
\renewcommand{\arraystretch}{1.12}
\begin{tabular}{@{}p{0.46\linewidth}p{0.50\linewidth}@{}}
\toprule
Setting & Value \\
\midrule
\multicolumn{2}{@{}l}{\textit{Large generative model}} \\
Pretrained backbone & Wan2.2-TI2V-5B \\
LoRA configuration & Rank $r=512$, scaling factor $\alpha=512$ \\
LR projection hidden widths & 768, 3,072 \\
Temporal attention window & 4 keyframes \\
Spatial attention window & $22\times40$ \\
Trainable parameters & 1.32B (LoRA and LR projection) \\
\midrule
\multicolumn{2}{@{}l}{\textit{Lightweight VSR network}} \\
Transformer & 10 blocks, width 512 \\
LR encoder channels & 64, 128, 128 \\
LR encoder strides & 2, 2, 1 \\
Reference input projection & $1\times1$ convolution, 48 to 512 channels \\
Video attention window & 2 frames, $24\times24$ spatially \\
Reference attention window & $48\times48$ for self-attention and reading \\
RGB output head & Width 1,024; PixelShuffle $\times16$ \\
Trainable parameters & 31.9M \\
\midrule
\multicolumn{2}{@{}l}{\textit{GAN discriminator}} \\
Wan feature layers & 7, 14, 21 (frozen) \\
Pooling head & 4 queries, 12 heads, width 3,072 \\
Trainable parameters & 198.4M (pooling head) \\
\bottomrule
\end{tabular}
\end{table}

\subsection{Base Training Settings}
\label{app:base_training_config}

\begin{table}[t]
\centering
\caption{\textbf{Base training settings.} Batch sizes are global. GAN learning rate and weight decay apply to both generator and discriminator.}
\label{tab:app_base_training}
\small
\setlength{\tabcolsep}{4pt}
\renewcommand{\arraystretch}{1.12}
\begin{tabular}{@{}p{0.28\linewidth}p{0.21\linewidth}p{0.21\linewidth}p{0.21\linewidth}@{}}
\toprule
Setting & Flow matching & One-step GAN & Lightweight VSR \\
\midrule
Initialization & Wan2.2 & Adapted generator & Random \\
Video frames per clip & 8 & 8 & 16 \\
HR crop & $704\times1280$ & $704\times1280$ & $1024\times1536$ \\
Global video / image batch & 32 / 96 & 32 / 96 & 64 / --- \\
Updates & 75K & 5K & 60K \\
GPUs & 16 & 16 & 16 \\
\midrule
Optimizer & AdamW & AdamW & AdamW \\
Learning rate & $2\times10^{-5}$ & $10^{-5}$ & $10^{-4}$ \\
Weight decay & $10^{-4}$ & $10^{-4}$ & 0 \\
Maximum gradient norm & 1 & G: 2; D: 5 & 1 \\
\bottomrule
\end{tabular}
\end{table}

All three runs use BF16 computation and FP32 gradient reduction.
For flow matching, AdamW uses $(\beta_1,\beta_2)=(0.9,0.95)$.
The learning rate warms up for 100 updates from 1\% of its target value and then stays constant.
Diffusion times follow a logit-normal distribution with mean 0, standard deviation 1, and shift 3; text-conditioning dropout is 0.1.

For one-step GAN distillation, the adversarial and feature-space R1 weights in Eq.~(\ref{eq:keyframe_training}) are 0.1 and 1,000.
The generator and discriminator use AdamW coefficients $(0.5,0.99)$ and $(0,0.99)$, respectively.
The discriminator trains alone for the first 20 updates, warming its learning rate from 1\% of the target value; the generator then trains at a constant learning rate.
Feature-space R1 is applied at every discriminator update, and the selected generator uses an EMA decay of 0.999.

The lightweight VSR network uses an all-frame loss of $L_1+0.1\,\operatorname{LPIPS}$, with AlexNet LPIPS~\citep{zhang2018lpips}.
The right reference is dropped with probability 0.25.
AdamW uses $(\beta_1,\beta_2)=(0.9,0.99)$ and $\epsilon=10^{-8}$; the learning rate warms up for 500 updates from 1\% of its target value and then stays constant.
The selected network uses an EMA decay of 0.9995.

\subsection{VARO Training Settings}
\label{app:varo_settings}

VARO trains on 46-frame clips with keyframes at positions 1, 16, 31, and 46, giving three complete intervals under dual-endpoint conditioning.
For each LR--HR clip, the sampling policy draws four candidate reference sequences, each of which receives one combined reward for its full rollout.
Two clips per update therefore produce eight scored rollouts.

We standardize each reward component and the system- and reference-level totals using fixed $z$-score statistics.
For each clip, we subtract the mean reward of its four candidates and divide by the global standard deviation across the update batch; clipping to $[-1,1]$ and rescaling gives the DiffusionNFT weight $r\in[0,1]$.
After each optimizer step, the sampling policy retains a fraction $\eta_i$ of its previous weights and takes $1-\eta_i$ from the updated generator, where $i$ counts VARO updates.
The level weights in Table~\ref{tab:app_varo_configuration} place greater emphasis on the emitted video while preserving direct feedback on decoded references.

\begin{table}[!h]
\centering
\caption{\textbf{VARO training settings.} Batch size counts LR--HR clips. Weight tuples follow the component order in Eq.~(\ref{eq:varo_rewards}).}
\label{tab:app_varo_configuration}
\small
\setlength{\tabcolsep}{5pt}
\renewcommand{\arraystretch}{1.06}
\begin{tabular}{@{}p{0.42\linewidth}p{0.54\linewidth}@{}}
\toprule
Setting & Value \\
\midrule
\multicolumn{2}{@{}l}{\textit{Rollout}} \\
Training clip & 46 frames at $704\times1280$ HR \\
Reference schedule & $\Delta=15$, dual-endpoint \\
Candidates per clip & 4 \\
\midrule
\multicolumn{2}{@{}l}{\textit{Optimization}} \\
Optimizer and learning rate & AdamW, $5\times10^{-6}$ \\
Batch size & 2 clips (8 rollouts) \\
Training updates & 2,000 \\
Hardware and precision & 16 GPUs, BF16 \\
Sampling-policy EMA & $\eta_i=\min(0.001i,0.5)$ \\
DiffusionNFT mixing coefficient $\beta$ & 1 \\
\midrule
\multicolumn{2}{@{}l}{\textit{Reward configuration}} \\
Video MUSIQ stride & 4 frames \\
System reward weights & $(0.40,0.25,0.25,0.10)$ \\
Reference reward weights & $(0.50,0.40,0.10)$ \\
Level weights $(\lambda_{\mathrm{sys}},\lambda_{\mathrm{ref}})$ & $(1.00,0.25)$ \\
\bottomrule
\end{tabular}
\end{table}
 
\section{Additional Results and Analysis}
\label{app:additional_results}

\subsection{Efficiency Across Resolutions and Reference Intervals}
\label{app:relay_inference_efficiency}

\paragraph{Comparison with baselines.}
Table~\ref{tab:app_baseline_resolution} extends the main-paper 1080p efficiency
comparison to four output resolutions. Each method processes 201 frames of
$4\times$ VSR on one NVIDIA A100-SXM4-80GB. FPS measures the complete sequence
of model CUDA-event calls after warmup; input/output handling, preprocessing,
and transfers outside those calls are excluded. Peak memory is PyTorch
allocated memory.
Spatial tiling is enabled for baselines when needed to avoid
out-of-memory errors.

\begin{table}[t]
\centering
\caption{\textbf{Inference efficiency across output resolutions.} Model FPS
and peak allocated memory (GB) for 201-frame $4\times$ VSR on one A100-80G.
RelayVSR uses $\Delta=15$. Best and second-best entries in each column are
bold and underlined, respectively.}
\label{tab:app_baseline_resolution}
\footnotesize
\setlength{\tabcolsep}{2pt}
\renewcommand{\arraystretch}{1.08}
\begin{tabular*}{\textwidth}{@{\extracolsep{\fill}}lrrrrrrrr@{}}
\toprule
\multirow{2}{*}{Method} &
\multicolumn{2}{c}{720p} &
\multicolumn{2}{c}{1080p} &
\multicolumn{2}{c}{1440p} &
\multicolumn{2}{c}{2160p} \\
\cmidrule(lr){2-3}\cmidrule(lr){4-5}\cmidrule(lr){6-7}\cmidrule(l){8-9}
& FPS $\uparrow$ & GB $\downarrow$
& FPS $\uparrow$ & GB $\downarrow$
& FPS $\uparrow$ & GB $\downarrow$
& FPS $\uparrow$ & GB $\downarrow$ \\
\midrule
RealBasicVSR & 41.758 & 11.04 & 18.502 & 24.81 & 10.671 & 38.04 & 4.888 & 66.54 \\
RealViformer & 31.884 & \textbf{5.44} & 14.873 & \textbf{12.36} & 8.689 & 21.66 & 3.880 & 39.03 \\
Stream-DiffVSR & 1.759 & \underline{10.24} & 0.849 & 25.92 & 0.468 & 61.67 & 0.169 & 44.39 \\
FlashVSR & 16.518 & 12.90 & 7.799 & 24.45 & 4.410 & 40.61 & 1.302 & 67.99 \\
SwiftVR & 51.056 & 21.25 & 23.483 & 31.36 & 13.367 & 40.74 & 5.773 & 65.26 \\
SeedVR2-3B & 2.645 & 75.05 & 0.788 & 73.89 & 0.438 & 74.88 & 0.145 & 61.46 \\
DOVE & 1.529 & 30.52 & 0.563 & 41.08 & 0.238 & 57.12 & 0.152 & 42.11 \\
SparkVSR & 1.525 & 30.62 & 0.564 & 41.30 & 0.237 & 57.51 & 0.152 & 42.34 \\
\midrule
\textbf{RelayVSR (dual-endpoint)} & \underline{58.320} & 11.92 & \underline{29.290} & 13.82 & \underline{17.590} & \underline{16.47} & \underline{8.230} & \underline{24.25} \\
\textbf{RelayVSR (first-keyframe)} & \textbf{68.590} & 11.86 & \textbf{35.890} & \underline{13.72} & \textbf{21.440} & \textbf{16.31} & \textbf{10.050} & \textbf{23.86} \\
\bottomrule
\end{tabular*}
\end{table}

Both RelayVSR modes have higher model FPS than the measured baselines at each
resolution. At 2160p, dual-endpoint conditioning reaches 8.23 FPS with
24.25\,GB peak allocated memory, compared with 1.302 FPS and 67.99\,GB for
FlashVSR and 5.773 FPS and 65.26\,GB for SwiftVR. First-keyframe
conditioning reaches 10.05 FPS with 23.86\,GB.

\paragraph{Reference-interval sweep.}
Table~\ref{tab:app_relay_efficiency} isolates the effect of keyframe spacing
on RelayVSR throughput. It uses the same 201-frame model-computation protocol,
with three timed runs per setting and both models included in each FPS value.

\begin{table}[t]
\centering
\caption{\textbf{RelayVSR throughput across reference intervals.} Model FPS
for 201-frame videos at four output resolutions on one A100-80G.}
\label{tab:app_relay_efficiency}
\small
\setlength{\tabcolsep}{5pt}
\renewcommand{\arraystretch}{1.08}
\begin{tabular*}{\textwidth}{@{\extracolsep{\fill}}llrrrr@{}}
\toprule
Mode & $\Delta$ & 720p & 1080p & 1440p & 2160p \\
\midrule
Dual-endpoint & 5 & 36.02 & 18.10 & 10.58 & 4.91 \\
              & 10 & 50.80 & 25.60 & 15.24 & 7.12 \\
              & 15 & 58.32 & 29.29 & 17.59 & 8.23 \\
              & 20 & 64.34 & 32.20 & 19.56 & 9.17 \\
              & 25 & 68.10 & 34.13 & 20.75 & 9.69 \\
\midrule
First-keyframe & 5 & 40.55 & 20.42 & 11.89 & 5.55 \\
               & 10 & 60.69 & 30.53 & 18.09 & 8.46 \\
               & 15 & 68.59 & 35.89 & 21.44 & 10.05 \\
               & 20 & 80.58 & 40.64 & 24.52 & 11.41 \\
               & 25 & 85.94 & 43.57 & 26.34 & 12.39 \\
\bottomrule
\end{tabular*}
\end{table}

Increasing $\Delta$ from 5 to 25 raises 2160p throughput from 4.91 to 9.69
FPS in dual-endpoint mode and from 5.55 to 12.39 FPS in first-keyframe mode.
At $\Delta=15$, first-output model time at 2160p is 1.249\,s for dual-endpoint
and 0.690\,s for first-keyframe conditioning; these times exclude input wait.

\subsection{Additional Qualitative Comparisons}
\label{app:additional_qualitative}

\begin{figure}[p]
    \centering
    \includegraphics[width=\textwidth]{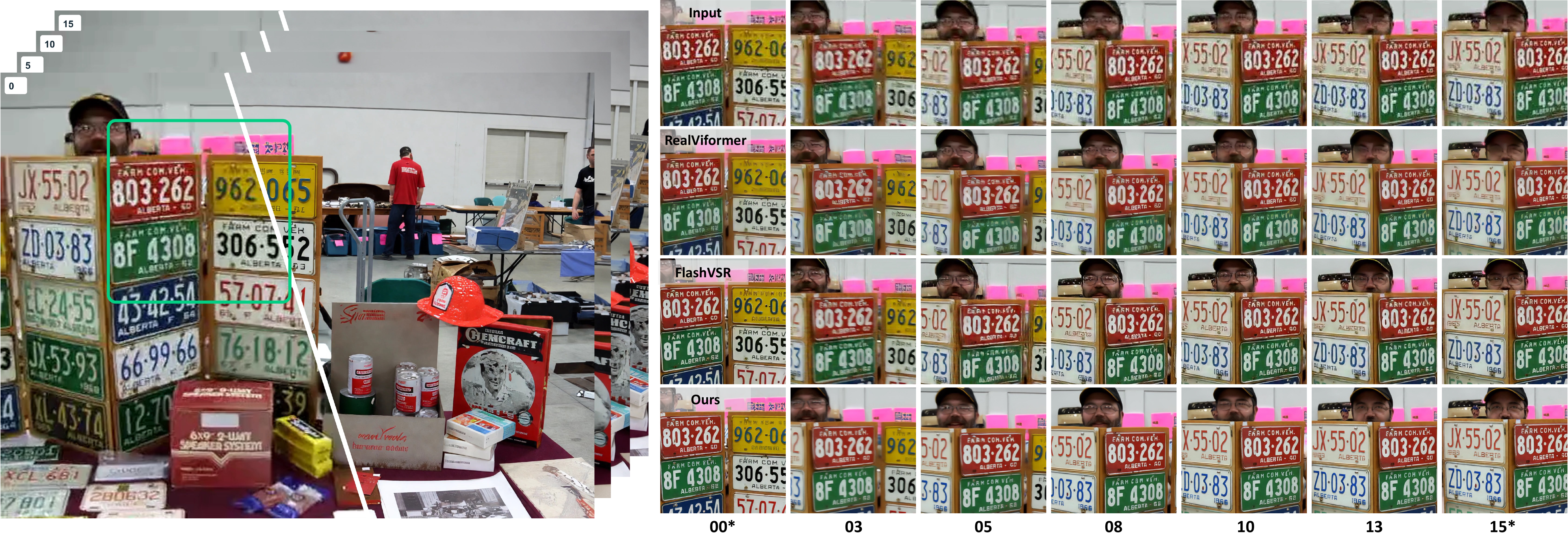}
    \caption{\textbf{Detail propagation between sparse references.}
    Left: the source frame locates the fixed crop. Right: keyframes 00 and 15
    (marked $*$) and five intervening frames compare the same license-plate
    region in the LR input, RealViformer, FlashVSR, and RelayVSR.}
    \label{fig:app_keyframe_propagation}
\end{figure}

Figure~\ref{fig:app_keyframe_propagation} follows a license-plate
crop between reference keyframes 00 and 15.
Using both endpoint references, RelayVSR retains recognizable
plate characters and colors at intermediate frames 03, 05, 08,
10, and 13 without generating references for these frames.

\begin{figure}[p]
    \centering
    \includegraphics[width=\textwidth]{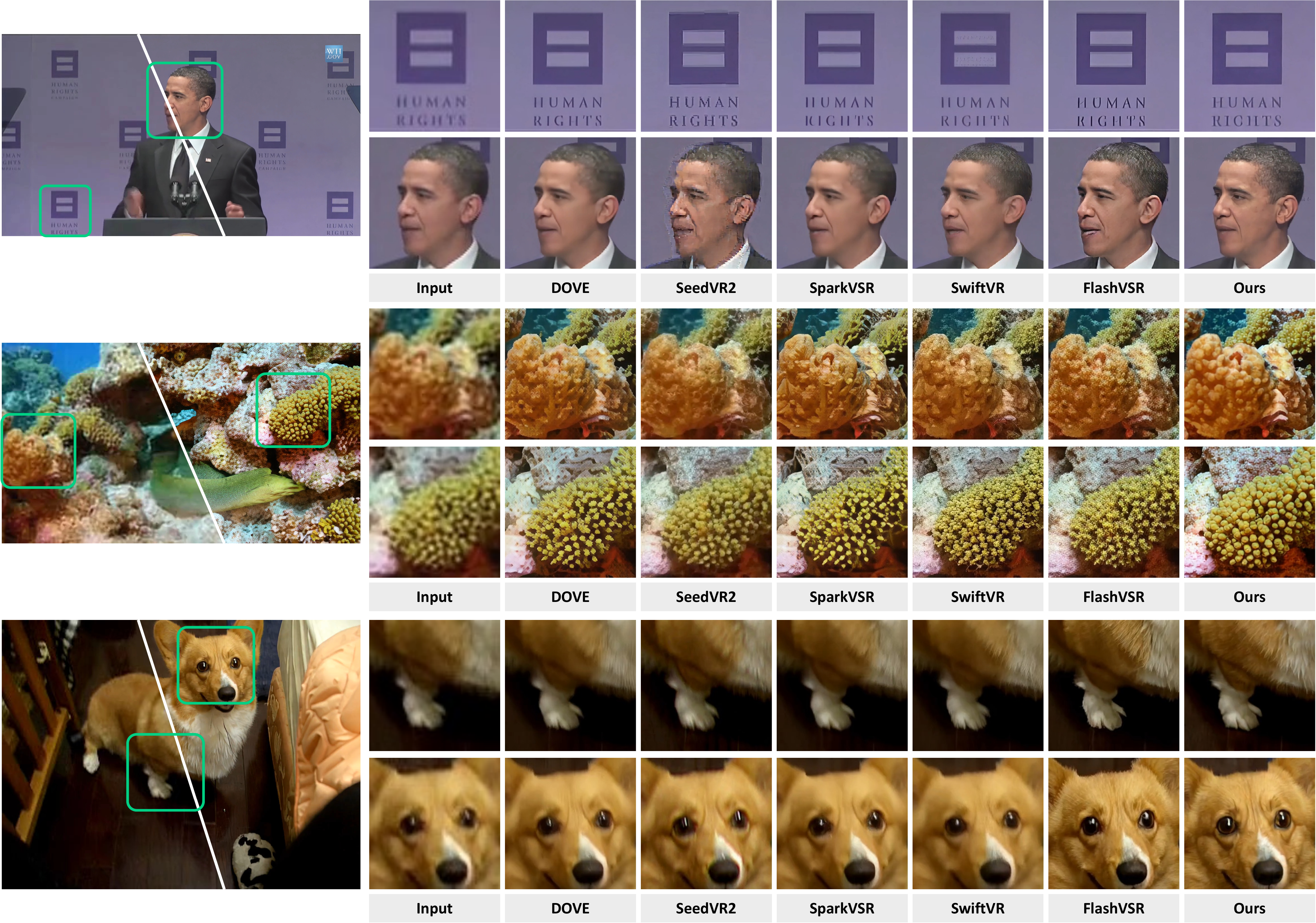}
    \caption{\textbf{Additional visual comparisons I.}
    Three examples highlight printed backdrop text and a face, underwater
    coral structure, and a dog's face and fur. Left: full-frame context with
    marked crops. Right: aligned crops from the input and six VSR methods,
    including RelayVSR.}
    \label{fig:app_compare2}
\end{figure}

\begin{figure}[p]
    \centering
    \includegraphics[width=\textwidth]{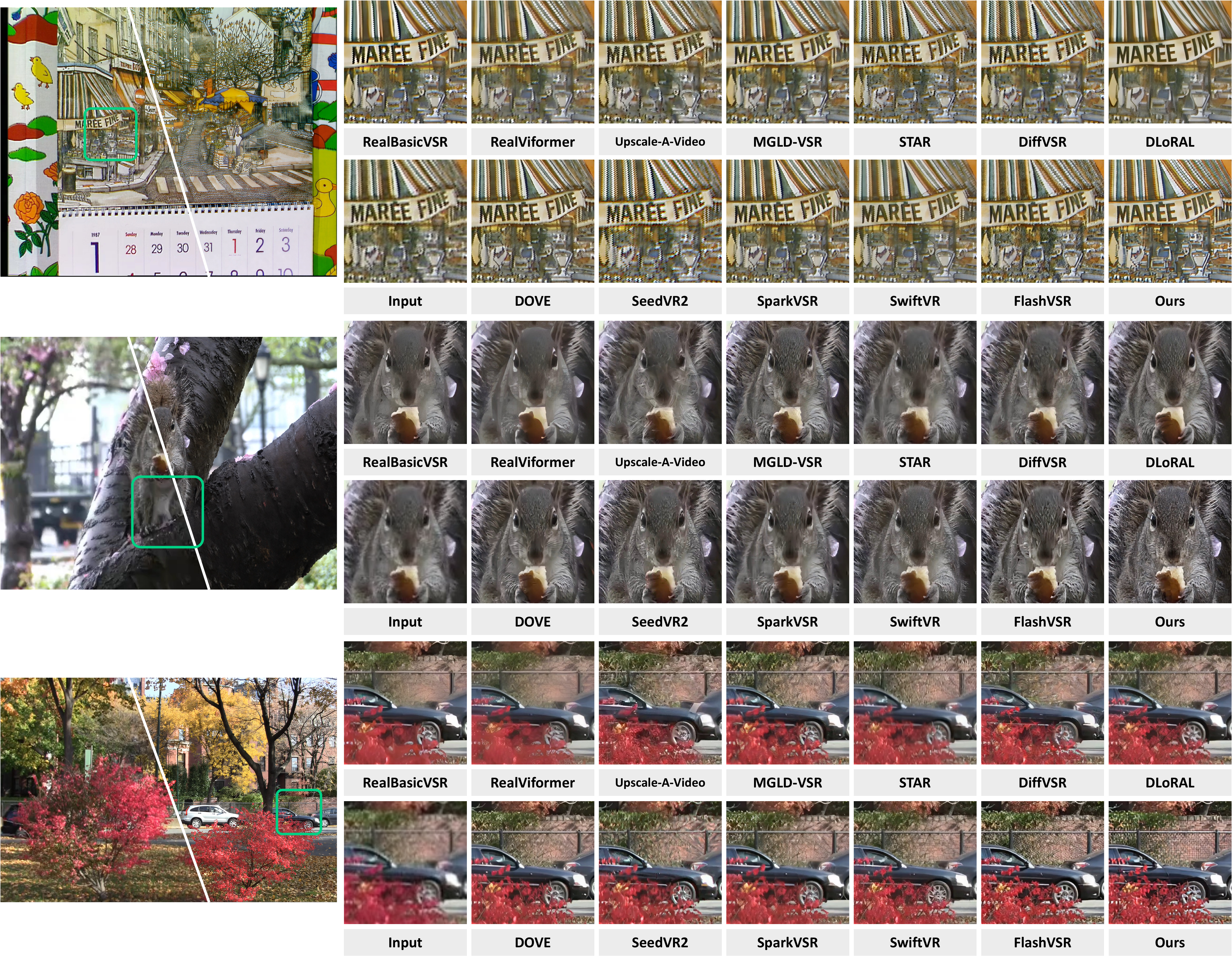}
    \caption{\textbf{Additional visual comparisons II.}
    Aligned crops show lettering on a street sign, a squirrel, and a car
    partly seen through foliage. Each example includes full-frame context
    at left and comparisons with earlier and recent VSR methods at right.}
    \label{fig:app_compare3}
\end{figure}

Figures~\ref{fig:app_compare2} and~\ref{fig:app_compare3} extend the
main-paper visual comparison to six further scenes. Their aligned crops show
printed characters, faces, coral, fur, and a car partly obscured by foliage.
These single-frame comparisons broaden the range of visible details, while
Fig.~\ref{fig:app_keyframe_propagation} shows how a detail persists between
reference updates.

\subsection{Long-Video Quality on LongVSR60}
\label{app:long_video_quality}

LongVSR60 extends evaluation to 30 real-world and 30 AI-generated
single-shot videos of approximately 1,000 frames each. In addition to
NIQE, MUSIQ, CLIP-IQA, and DOVER, we report subject consistency (SC),
background consistency (BC), and motion smoothness (MS) following
VBench~\citep{huang2024vbench} as supplementary temporal indicators.
SC and BC measure cross-frame appearance consistency using
DINO and CLIP features, respectively.
Motion smoothness (MS) follows the interpolation-based criterion
in VBench, measuring agreement between observed frames and
frames interpolated from temporal neighbors.
Table~\ref{tab:longvsr60_quality} reports RelayVSR
and selected baselines on the 60-video benchmark.

\begin{table}[t]
\centering
\caption{\textbf{LongVSR60 quality.} The benchmark contains 30 real-world
and 30 AI-generated videos. NIQE, MUSIQ, and CLIP-IQA use 100 uniformly
sampled frames per video; SC, BC, and MS are supplementary temporal
indicators. Best and second-best values are bold and underlined, respectively.}
\label{tab:longvsr60_quality}
\small
\setlength{\tabcolsep}{4pt}
\renewcommand{\arraystretch}{1.12}
\begin{tabular*}{\textwidth}{@{\extracolsep{\fill}}lrrrrrrr@{}}
\toprule
& \multicolumn{4}{c}{No-reference quality}
& \multicolumn{3}{c}{Temporal indicators} \\
\cmidrule(lr){2-5}\cmidrule(lr){6-8}
Method & NIQE $\downarrow$ & MUSIQ $\uparrow$ &
CLIP-IQA $\uparrow$ & DOVER $\uparrow$ &
SC $\uparrow$ & BC $\uparrow$ & MS $\uparrow$ \\
\midrule
RealViformer & 5.0770 & 46.17 & 0.3819 & 0.5053 & \textbf{0.8846} & \textbf{0.9198} & \textbf{0.9848} \\
Stream-DiffVSR & \underline{4.0491} & 54.52 & 0.4694 & 0.5704 & 0.8830 & 0.9161 & 0.9813 \\
SwiftVR & 4.2822 & 53.89 & 0.4640 & 0.5512 & 0.8831 & \underline{0.9177} & 0.9824 \\
FlashVSR & \textbf{3.8160} & \textbf{55.15} & \textbf{0.4747} & \underline{0.5895} & 0.8829 & 0.9144 & 0.9802 \\
\midrule
RelayVSR & 4.1396 & \underline{54.72} & \underline{0.4701} & \textbf{0.5911} & \underline{0.8837} & 0.9157 & \underline{0.9827} \\
\bottomrule
\end{tabular*}
\end{table}
 
\subsection{User Study}
\label{app:user_study}

We conducted a blind pairwise study with 12 participants and 36 videos:
12 from VideoLQ and 12 each from the real-world and AI-generated subsets
of LongVSR60. We use 10--15\,s excerpts, or full videos when shorter;
LongVSR60 excerpts are taken after full-sequence inference.
RelayVSR uses dual-endpoint conditioning with $\Delta=15$ and is compared
with DOVE, SeedVR2, SparkVSR, SwiftVR, and FlashVSR-Tiny under the main
evaluation settings.

The LR input and anonymized A/B outputs were played synchronously at
matched resolution and frame rate, with randomized left--right placement
and trial order. Participants judged
overall visual quality, fidelity to observable LR content, and temporal
stability, choosing A better, similar, or B better. Unable-to-judge fidelity
responses were excluded from that criterion's score.

Following the GSB protocol~\citep{zhuang2026flashvsr}, we report
$100\%\times(G-B)/(G+S+B)$, where $G$, $S$, and $B$ count better, similar,
and worse judgments for RelayVSR. Results are reported in
Table~\ref{tab:app_user_study}.

\begin{table}[t]
\centering
\caption{\textbf{User study.} GSB scores (\%) compare RelayVSR with each baseline.
Positive scores favor RelayVSR; negative scores favor the baseline.}
\label{tab:app_user_study}
\small
\setlength{\tabcolsep}{5pt}
\begin{tabular*}{\textwidth}{@{\extracolsep{\fill}}lccc@{}}
\toprule
Baseline & Overall quality $\uparrow$ & Content fidelity $\uparrow$ & Temporal stability $\uparrow$ \\
\midrule
DOVE & +53.6 & -8.9 & -8.7 \\
SeedVR2 & +46.8 & +13.5 & +13.1 \\
SparkVSR & +43.9 & +21.9 & +5.9 \\
SwiftVR & +26.7 & +13.2 & +11.9 \\
FlashVSR & -3.3 & +9.0 & +17.0 \\
\bottomrule
\end{tabular*}
\end{table}

RelayVSR receives higher overall-quality preference than DOVE, SeedVR2,
SparkVSR, and SwiftVR, with GSB scores ranging from +26.7\% to +53.6\%.
Against FlashVSR, the near-neutral overall-quality score (-3.3\%) supports
comparable perceived quality, while content fidelity (+9.0\%) and temporal
stability (+17.0\%) favor RelayVSR. These perceptual results complement the
model-computation efficiency gains reported in
Sec.~\ref{sec:efficiency_streaming}.

\section{Limitations and Future Work}
\label{app:limitations}

RelayVSR uses a fixed keyframe interval, while dual-endpoint conditioning
requires bounded lookahead. Adapting keyframe placement and reference
selection to content changes and latency budgets could improve the balance
between video quality and computational cost.
Texture or color errors in generated references can propagate across output
frames. Complementing VARO with reference reliability estimation at inference
could guide memory reading and reference refresh, helping the lightweight VSR
network adjust its reliance on generated details.
Although sparse invocation reduces average computation, the large generative
model still contributes to memory use. Our efficiency results are measured
on an NVIDIA A100. Model compression and smaller reference generators could
support deployment on devices with tighter resource budgets.

\clearpage

\clearpage

\end{document}